\def\doublecolumn{1} % double column = 1, single column = 0
\if 1\doublecolumn
  \documentclass[journal]{IEEEtran}
\else
  \documentclass[12pt, draftclsnofoot, onecolumn, letterpaper]{IEEEtran}
\fi

\def\blind{1} % d-blind = 0 s-blind = 1

\usepackage[latin9]{inputenc}
\usepackage{algpseudocode}
\usepackage{algorithm}
\usepackage{array}
\usepackage{makecell} 
\usepackage{float}
\usepackage{mathtools}
\usepackage{amsmath}
\usepackage{amsfonts}
\usepackage{amsthm}
\usepackage{amssymb}
\usepackage{graphicx}
\usepackage{wasysym}
\usepackage{setspace}
\usepackage{color}
\usepackage{bm}
\usepackage{cases}
\usepackage{tikz}
\usetikzlibrary{calc}
\usepackage{flowchart}
\usepackage{bbm}
\usepackage{adjustbox}
\usepackage{hhline}
\usepackage{makecell}

\usepackage{epsfig}
\usepackage{cite}
\usepackage{graphicx}
\usepackage{lipsum}
\usepackage{multirow}
\usepackage{array}
\usepackage{enumerate}
\usepackage{enumitem}
\usepackage{graphicx}
\usepackage{times}
\usepackage[framemethod=tikz]{mdframed}
\definecolor{cccolor}{rgb}{1,1,1}

\usepackage{subfig}
\usepackage{caption}
\newtheorem{theorem}{Theorem}
\newtheorem{lemma}{Lemma}

\newtheorem{proposition}{Proposition}
\newtheorem{definition}{Definition}
\newtheorem{assumption}{Assumption}

\usepackage{multicol,multirow,adjustbox}
\usepackage{booktabs}

\newcommand{\ie}[0]{\textit{i.e.}}
\newcommand{\etc}[0]{\textit{etc}}

\usepackage[capitalize]{cleveref}
\crefname{section}{Sec.}{Secs.}
\Crefname{section}{Section}{Sections}
\crefname{table}{Tab.}{Tabs.}
\Crefname{table}{Table}{Tables}
\crefname{assumption}{Assumption}{Assumptions}

\makeatletter
\def\@IEEEsectpunct{.\ \,}
\def\paragraph{\@startsection{paragraph}{4}{\z@}{1.5ex plus 1.5ex minus 0.5ex}%
{0ex}{\normalfont\normalsize\bfseries}}
\makeatother

\usepackage{pifont,xspace}
\newcommand{\cmark}{\ding{51}\xspace}%
\newcommand{\xmarkg}{\textcolor{lightgray}{\ding{55}}\xspace}%

\begin{document}

\title{
    Rethinking Model Inversion Attacks  With Patch-Wise Reconstruction
}
\if 1\blind
\author{Jonggyu Jang,~\IEEEmembership{Member,~IEEE}, Hyeonsu Lyu,~\IEEEmembership{Student Member,~IEEE}, and  Hyun~Jong~Yang,~\IEEEmembership{Member,~IEEE}
    \thanks{
    J. Jang is with Institute of New Media and Communications, Seoul National University, Seoul 08826, Republic of Korea, (email: jgjang@snu.ac.kr).
    H. Lyu and is with Department of Electrical Engineering, Pohang University of Science and Technology (POSTECH), Pohang 37673, Republic of Korea, (e-mail: hslyu4@postech.ac.kr). 
    H. J. Yang (corresponding author) is with Department of Electrical and Computer Engineering, Seoul National University, Seoul 08826, Republic of Korea, (email: hjyang@snu.ac.kr).
    }
}
\else
\author{Anonymous Submission
    }
\fi

\maketitle

\begin{abstract}
Model inversion (MI) attacks aim to infer or reconstruct the training dataset through reverse-engineering from the target model's weights.
Recently, significant advancements in generative models have enabled MI attacks to overcome challenges in producing photo-realistic replicas of the training dataset, a technique known as \textbf{generative MI}.
The generative MI primarily focuses on identifying latent vectors that correspond to specific target labels, leveraging a generative model trained with an auxiliary dataset. 
However, an important aspect is often overlooked: the MI attacks fail if the pre-trained generative model lacks the coverage to create an image corresponding to the target label, especially when \textbf{there is a significant difference between the target and auxiliary datasets}.
To address this gap, we propose the \textbf{Patch-MI} method, inspired by a \textbf{jigsaw puzzle}, which offers a novel probabilistic interpretation of MI attacks.
Even with a dissimilar auxiliary dataset, our method effectively creates images that closely mimic the distribution of image patches in the target dataset by patch-based reconstruction.
Moreover, we numerically demonstrate that the Patch-MI improves Top 1 attack accuracy by \textbf{5\%p} compared to existing methods.
\end{abstract}


\section{Introduction\label{sec:intro}}

The last decade has seen unparalleled advancements in deep learning technology, where this advancement has also led to a significant side effect: an increase in attempts to train models with private data.
According to~\cite{Song2017-pa, Feldman20_memory}, deep learning models are capable of retaining confidential data used during their training.
In essence, if the trained model weights are exposed to attackers, this poses a substantial risk for the leakage of private information.

The societal benefit of advancements in privacy attacks lies in the stimulation of the development of robust defenses, thereby enhancing data privacy~\cite{guo2019peid,kim2023optimized}.
% At this point, what is the societal benefit of advancements in privacy attacks?
% The answer is that privacy attacks can lead to the development of corresponding defenses, thereby reinforcing data privacy. 
There have been various types of neural network attacks in machine learning:  model inversion~\cite{Fredrikson14,Liu_TIFS,Zhu_TIFS,Zhang_TIFS}, weight reconstruction~\cite{Mili_2019}, domain inference~\cite{Gu_Chen_2023}, membership inference~\cite{shokri2017membership}, and backdoor attacks~\cite{liu2020reflection}.
Among these attack methods, model inversion (MI) attack is a strong and representative category of privacy attacks, which directly generates a replica of the training dataset used in the target deep learning model~\cite{Fredrikson14,Fredrikson15,Hidano17,knoblauch2019generalized,zhang2020secret,wang2021variational}.
However, as in a recent study~\cite{yu2024generator}, if there is no auxiliary dataset with a distribution similar to the target dataset, it is hard to generate data close to the target dataset.

\paragraph*{Problem setting}
In model inversion (MI) attacks, the availability of the target neural network's weights to the attacker distinguishes between a white box scenario (where weights are available) and a black box scenario (where they are not). 
Our study focuses on the white-box attack setting.
Let us consider a confidential dataset $\mathcal{D}_\text{tar}$, typically used in general image classification tasks.
That is, the dataset $\mathcal{D}_\text{tar}$ comprises image instances $\mathbf{x}\in\mathbb{R}^{c\times w\times h}$ (where $w$ and $h$ are the width and height of the images, respectively) and corresponding label instances $y\in\mathcal{C}=\{0,...,C-1\}$, where $C$ is the number of classes within the dataset. 
% In the threat model, a programmer trains a deep neural network (DNN) classifier using the dataset and subsequently shares the trained weights by uploading them to public platforms. 
With access to the model weights, a malicious attacker aims to approximate the distribution of the dataset inherent to the target classifier, denoted as $p_\text{tar}(\mathbf{x}|y)$. 
To articulate the approximation of the target posterior distribution (target classifier), we employ the following notation:
\begin{equation}
    \hat{p}_{\textrm{tar}}(y|\mathbf{x}): \mathbb{R}^{c\times w\times h} \rightarrow \Delta^{C},
\end{equation}
where $\Delta^{C}$ is the $(C-1)$-dimensional probability simplex.
In the next part, we will introduce representative MI attack methods. 
We note that more comprehensive literature reviews can be found in \cref{sec:related_works}.

\paragraph*{Baseline MI attack (BMI)}
In the initial phase of MI attacks, the baseline MI (BMI) attacks were proposed targeting tabular data classifiers, which are inherently simpler and less complex than classifiers used for high-dimensional data such as images, text, and voices~\cite{Fredrikson14, Fredrikson15, Hidano17}. 
In this context, we can formulate the objective function of the BMI as finding a data instance $\mathbf{x}\in\mathbb{R}^{c\times w\times h}$ that maximizes the target posterior probability as follows:
\begin{equation}\label{eq:BMI}
    \mathbf{x}^* = \arg\min_{\mathbf{x}\in\mathbb{R}^{c\times w\times h}} -\log \hat{p}_\text{tar}(y|\mathbf{x}),
\end{equation}
where this loss function is referred as \textit{identity loss} in later studies~\cite{zhang2020secret,Nguyen_2023_CVPR}.
In Equation \eqref{eq:BMI}, the target image $\mathbf{x}^*$ can be easily found by applying gradient descent, \ie, $\frac{\partial \log p_\text{tar}(y|\mathbf{x})}{\partial \mathbf{x}}$.
However, the lack of photo-realism can diminish the practical utility of the reconstructed images, as they might not be sufficiently detailed or accurate for high-dimensional datasets.

\paragraph*{Generative MI attack (GMI)}

To overcome the challenge of producing unnatural images in BMI, several studies have explored the use of generative models, which are pre-trained with an auxiliary dataset.
We denote the pre-trained generative model as $G$ and its corresponding discriminator as $D$. 
In generative MI (GMI), the attackers seek to find an optimal latent vector $\mathbf{z}^*$ such that when passed through $G$, it maximizes the posterior probability $\hat{p}_\mathrm{tar}(y|G(\mathbf{z}))$ for given class $y$~\cite{yang2019adversarial,Yang_19_GANMI,zhang2020secret}.
Moreover, an additional term for maximizing the discriminator's output is used for enhancing photo-realism.
That is, the objective function can be written as
\begin{equation}\label{eq:GMI}
\begin{split}
\mathbf{z}^*= \arg\max_{\mathbf{z}} \lambda \log \hat{p}_\mathrm{tar}(y|&(G (\mathbf{z})) + \log(\sigma( D (G (\mathbf{z})))),
\end{split}
\end{equation}
where $\sigma(\cdot)$ denotes the Sigmoid function, and $\lambda>0$ is a weight on the identity loss. 
Several variations of the GMI have been proposed to enhance its effectiveness and tackle different challenges. These include: i) advanced identity loss~\cite{Nguyen_2023_CVPR}, ii) reinforcement learning~\cite{han2023reinforcement}, iii) supervised inversion~\cite{Tian_23_roleof}, and iv) adversarial examples~\cite{Zhou_23_boosting_MI_adv}.

\paragraph*{Variational MI attack (VMI)} 
In the variational MI (VMI) attack~\cite{wang2021variational}, a unique approach based on variational inference is used to avoid collapsing the output images.
We denote $q(\mathbf{z})$ as the distribution of latent vector $\mathbf{z}$.
Then, the objective function is derived from the Kullback-Leibler (KL) divergence between target/replica data distribution:
\begin{equation}
\begin{split}
    q^*(\mathbf{z})  = \arg\min_{q\in Q_\mathbf{z}} \big\{\mathbb{E}_{\mathbf{z}\sim q}\big[ D_\mathrm{KL}&(q(\mathbf{z})\Vert  \hat{p}_\mathrm{tar}(\mathbf{z})) 
  - \lambda \log \hat{p}_\mathrm{tar}(\mathbf{z})\big] \big\}.
\end{split}
\end{equation}
% In the VMI, StyleGAN2 \cite{Karras2019stylegan2} and normalizing flows \cite{pmlr-v37-rezende15} are used for better disentanglement. 
Unlike GMI, where the goal is to find a specific latent vector $z$ that best represents a given class $y$, VMI seeks to approximate the entire distribution of $z$.

\begin{figure*}[t]
    \centering
    \includegraphics[width=0.99\linewidth]{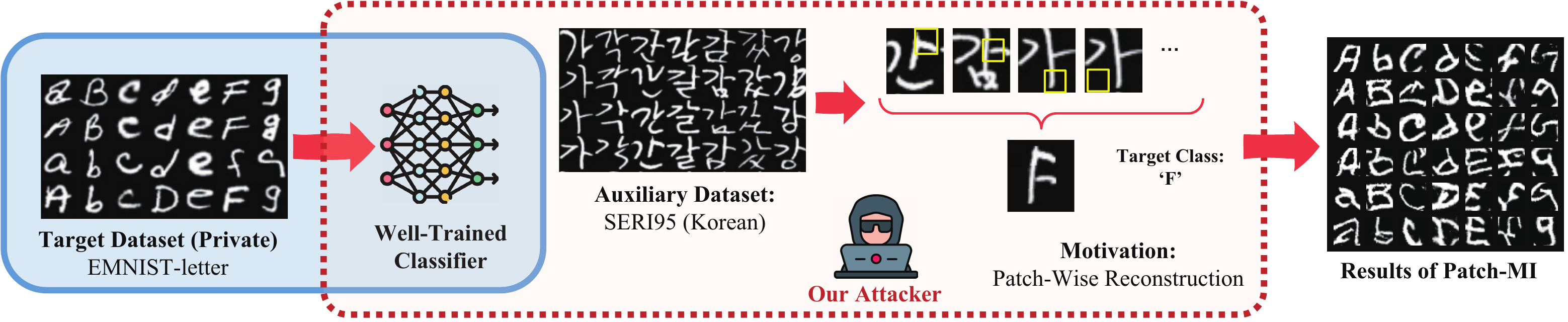}
    \caption{
    Illustration of our motivation. Using an 8x8 patch-wise discriminator, the Patch-MI method successfully synthesizes target images (English alphabet) from the auxiliary dataset (Korean Hangeul), even though the auxiliary dataset is dissimilar to the Alphabet dataset.
    }
    \label{fig:motivation_1}
\end{figure*}

\paragraph*{Main challenge}
In this paper, we mainly tackle the challenge often overlooked in existing MI methods: the \textit{statistical dissimilarity} between target and auxiliary datasets.
Prior works, including VMI, GMI, and their variations, often rely on the strong assumption that attackers have access to either \textit{a subset of the target dataset} or \textit{partially overlapped datasets}.
For instance, many studies targeting MNIST classifiers assume access to \textit{a part of the MNIST dataset}~\cite{Zhou_23_boosting_MI_adv,zhang2020secret}. 
Alternatively, other studies utilize the EMNIST dataset as an auxiliary dataset, which is \textit{partially overlapped} with the MNIST dataset~\cite{wang2021variational,yang2019adversarial,Yang_19_GANMI}\footnote{Digits $\{0,1,2,4,5,7,9\}$ and alphabets $\{O,I,Z,A,S,t,g\}$.}, where \textbf{the pre-trained generator cannot generate images of the non-overlapping classes}. 
However, such access is highly unlikely, making these assumptions impractical.
Furthermore, the results of the previous works demonstrate that the attackers often fail in the non-overlapped classes~\cite{yang2019adversarial,wang2021variational,zhang2020secret}.

\paragraph*{Intuition and Contribution}
In response to the challenge of dissimilarity between target and auxiliary datasets, we propose Patch-MI, a novel method for model inversion attacks, the motivation of which is illustrated in \cref{fig:motivation_1}. 
As depicted in the figure, our proposed method is inspired by jigsaw puzzle, combining image patches for generating a desired image. 
By doing this, this method is particularly effective when the target and auxiliary datasets have no overlapping classes. 
% the Patch-MI is inspired by the concept of combining partial characteristics from an auxiliary dataset to reconstruct the training dataset of a target classifier.
% This method is particularly effective when the target and auxiliary datasets have no overlapping classes.
To this end, we have the following contributions:
% We postulate a publicly available classifier trained with a private dataset and demonstrate that our patch generative adversarial network (GAN)-inspired technique effectively rebuilds the training dataset despite dissimilar distribution with the auxiliary dataset. Utilizing a GAN-like approach with a patch-based discriminator, we enumerate the key contributions as follows:
\begin{itemize}
    \item Our approach introduces a novel probabilistic perspective to MI attacks, which allows for a more nuanced understanding of how these attacks can be structured and optimized.
    We have developed a loss function specifically aimed at minimizing the Jensen-Shannon (JS) divergence between the confidential dataset and generated replicas.
    \item We employ a random transformation before forwarding generated images to the target classifier, with the aim of harmonic identity loss minimization, thereby allowing us to enhance attack accuracy.
    \item We conduct a benchmark using three different datasets. This demonstrates that our approach depends less on auxiliary datasets, especially in the context of datasets related to the shape of handwriting.
\end{itemize}

\section{Background and Related Works}\label{sec:related_works}

Here, we provide an in-depth analysis of MI attacks, whose objective is to uncover the sensitive attributes embedded within private datasets through target DNNs, assuming that attackers have access to these models.

\paragraph*{Optimization-based MI attack}
The concept of the MI attack was first delineated by \cite{Fredrikson14} within the context of logistic regression. This seminal work demonstrated that an attacker, armed with partial information about the target DNN, could indeed extract sensitive information. Building upon this foundation, subsequent studies \cite{Fredrikson15,Hidano17} formulated more generalized MI attack algorithms. By employing gradient descent, these enhanced approaches aimed to unveil sensitive information embedded within target models. However, such MI attacks are less effective against intricate and deep multi-layer neural networks. This limitation arises from the fact that data trained in the target classifier often occupy only a small portion of the image space, as elucidated in \cite{szegedy2015going}. A recent study introduces an MI attack technique that eschews the use of generative models for binary classifiers; yet, the challenge concerning photo-realistic images persists \cite{haim2022reconstructing}.

\paragraph*{Generative MI attack}
To rectify the limitations of optimization-based MI attacks, recent innovations have adopted generative models such as GANs \cite{goodfellow2020generative} and VAEs \cite{kingma2019introduction} to guarantee the photorealism of resultant images. In this context, the authors of \cite{zhang2020secret} skillfully employed a GAN pre-trained on public data for MI attacks, utilizing blurred or masked faces as supplementary information. 
Early influential works revealed that optimizing latent vectors to maximize posterior probability within target classifiers could expose sensitive traits of private datasets while maintaining photorealistic quality \cite{yang2019adversarial, Yang_19_GANMI, Khosravy_face_2020}. This marked a shift towards a generative MI paradigm, finely tuning latent space to minimize the target classifier's cross-entropy. Furthermore, the authors of \cite{chen2021knowledge} enhanced this approach by integrating soft labels from the target classifier, representing a significant advance in the development of refined and effective attack strategies.

\paragraph*{Extensions of generative MI attack}
In an effort to reconstruct the target class distribution, the authors of \cite{wang2021variational} pioneered a variational MI attack to ascertain the mean and variance of latent vectors, utilizing a deep flow model for accurate approximation. Subsequent contributions in \cite{pmlr-v162-struppek22a, han2023reinforcement, yuan2023pseudo, takahashi2023breaching} provided a robust enhancement through image transformation, reinforcement learning, and label-guidance. 
Additionally, the authors of \cite{wang2022reconstructing} investigated GAN-based MI attacks against ensemble classifiers, thereby evidencing enhanced dependability in the attack methodology.

\paragraph*{Remaining challenges}
Prevailing research frequently depends on the unfeasible supposition that the image corresponds with the target prior in the training of generative models, a misalignment potentially leading to inaccuracies. This paper contests this traditional notion by suggesting mitigation of this assumption, positing uniformity among the priors for various image segments. Our investigation focuses on enhancing MI attacks on DNN classifiers, especially in instances where the auxiliary and target datasets have dissimilar distributions.

\section{Patch Model Inversion Attack} 

In this section, we introduce a novel probabilistic interpretation of MI attacks. 
To this end, we formulate an optimization problem with the primary goal of minimizing the JS divergence between the attacker's distribution and the target data distribution. 
Furthermore, our utilization of a patch-wise discriminator negates the conventional assumption that the target and auxiliary datasets have identical distributions.
% This enhances the practicality and effectiveness of our method in real-world MI scenarios, where such an assumption may not necessarily be valid.

%  The attacker's objective is to uncover the sensitive characteristics inherent to the target classifier, the weights of which are readily accessible to the public. 
% Additionally, it is presumed that the attacker is capable of inferring the rough dataset type by examining the model's description, examples of which might include categories such as \textit{gray-scale handwriting} and \textit{color natural images}.

\subsection{Patch-MI: A New Probabilistic Interpretation}

In MI attacks, the attacker's objective is to uncover the training dataset's distribution inherent to the target classifier.
In this work, we assume that the target classifier is well-trained with the target dataset.
That is, as in \cref{assp:good_classifier}, we can approximate the target posterior probability $p_\text{tar}$ by the target classifier $\hat{p}_\text{tar}$.

\begin{assumption}[Well-trained classifier]\label{assp:good_classifier}
    We proceed under the assumption that the target classifier is robust and well-trained, capable of successfully categorizing the majority of the training data, \ie $\hat{p}_{\text{tar}}(y|\mathbf{x}) \approx p_{\text{tar}}(y|\mathbf{x}),~\forall (\mathbf{x},y)\in\mathcal{D}_\text{tar}$.
\end{assumption}

\paragraph*{Problem formulation}

We present an optimization problem that aims to minimize the JS divergence between the target data distribution  $\hat{p}_{\textrm{tar}}(\mathbf{x}|y)$ and the MI attacker $q(\mathbf{x})$, a kind of generative model.
This approach seeks to closely mimic the approximated target data distribution\footnote{In contrast to the utilization of the KL divergence as seen in \cite{wang2021variational}, we opt for the JS divergence as our objective function in this context. The preference for the JS divergence over the KL divergence is due to its ability to address issues related to asymmetry and the handling of non-overlapping distributions, as elucidated in \cite{huszar2015not}.
}:
\begin{equation}\label{eq:optimization_problem}
    \arg\min_{q \sim Q_\mathbf{x}}  D_{\text{JS}}\left(q(\mathbf{x}) \Vert \hat{p}_{\mathrm{tar}}(\mathbf{x}|y)\right),
\end{equation}
where $Q_\mathbf{x}$ denotes a set of possible probability distribution functions on a space $\mathbb{R}^{c\times w\times h}$.

Furthermore, to generate photo-realistic images, it is essential to utilize an auxiliary dataset encapsulating partial information akin to that of the target dataset, denoted by the distribution $p_\textrm{aux}(\mathbf{x})$.
As it is impossible to sample or acquire the approximated distribution $\hat{p}_\text{tar}(\mathbf{x}|y)$, our objective is transformed into obtaining and minimizing an upper bound of the problem articulated in Equation \eqref{eq:optimization_problem}. 
To this end, we have the following natural definition.
% \begin{assumption}
% \label{assp:equal_dist}
% The target dataset has an equal number of samples for each class, i.e., $p_\text{tar}(y) = q(y) = \frac{1}{C}$ for all $y\in\{0,1,..,C-1\}$.
% \end{assumption}

\begin{definition}[Sampling probability inequality]
\label{assp:ineq_during_training}
For an MI attacker $q(\mathbf{x})$ corresponding to the target class $y\in\{0, 1, ..., C-1\}$, the following inequality holds: $1=q(y|\mathbf{x})\ge p_\text{tar}(y|\mathbf{x})$. Similarly, $1=p_\mathrm{tar}(y|\mathbf{x})$ holds if $\mathbf{x}\sim p_\mathrm{tar}(\mathbf{x}|y)$. 
\end{definition}

To prove our main theorem, we have the following two Lemmas. 
% For the proofs of the Lemmas, please see Appendices \ref{subsec:prof_lem1} and \ref{subsec:prof_lem2}.
\begin{lemma}\label{lem:1}
By using \cref{assp:ineq_during_training}, if $\mathbf{x}\sim q(\mathbf{x})$, the following inequality is satisfied:
\begin{align}
    & \log \left( \frac{2q(\mathbf{x})}{q(\mathbf{x}) + \hat{p}_\mathrm{tar}(\mathbf{x}|y)} \right)   \\
    &~~~~~~ \le \log \left( \frac{2q(\mathbf{x})}{q(\mathbf{x}) + \hat{p}_\mathrm{tar}(\mathbf{x})} \right) -\frac{1}{2}\log \hat{p}_\mathrm{tar} (y|\mathbf{x})\nonumber. 
\end{align}
\begin{IEEEproof}
    In \cref{assp:ineq_during_training}, we have $1=q(y|\mathbf{x})\ge \hat{p}_\mathrm{tar}(y|\mathbf{x})$ if $\mathbf{x}\sim q(\mathbf{x})$ for the target class $y$. 
Also, since the target label is fixed as $y$ in an attack attempt, we let the target label probability as $p(c)$ is 1 if $c=y$ and 0 otherwise.\footnote{In this case, the following equality holds: $q(\mathbf{x}=q(\mathbf{x}|y)$.}
Then, the following inequality holds:
\begin{align}
    & \log \left( \frac{2q(\mathbf{x})}{q(\mathbf{x}) + \hat{p}_\mathrm{tar}(\mathbf{x}|y)} \right) \nonumber\\ 
    & = \log\left( \frac{2q(\mathbf{x})}{q(\mathbf{x}) + \hat{p}_\mathrm{tar}(\mathbf{x})\hat{p}_\mathrm{tar}(y|\mathbf{x})} \right) \nonumber\\ 
    & = \log \left( \frac{2q(\mathbf{x})}{q(\mathbf{x}) + \hat{p}_\mathrm{tar}(\mathbf{x})} \right) \\ 
        & ~~~~~~+ \log \left( \frac{q(\mathbf{x}) + \hat{p}_\mathrm{tar}(\mathbf{x})}{q(\mathbf{x}) + \hat{p}_\mathrm{tar}(\mathbf{x})\hat{p}_\mathrm{tar}(y|\mathbf{x})} \right)\nonumber \\
    & \le  \log \left( \frac{2q(\mathbf{x})}{q(\mathbf{x})  +\hat{p}_\mathrm{tar}(\mathbf{x})} \right) - \log \hat{p}_\mathrm{tar}(y|\mathbf{x}).\nonumber
\end{align}
\end{IEEEproof}
% \subsection{Proof of Lemma \ref{lem:1} }\label{subsec:prof_lem1}

\end{lemma}

\begin{lemma}\label{lem:2}
With \cref{assp:ineq_during_training}, if $\mathbf{x}\sim \hat{p}_\mathrm{tar}(\mathbf{x}|y)$, the following equality holds:
\begin{align}
    \log \left( \frac{2\hat{p}_\mathrm{tar}(\mathbf{x}|y)}{q(\mathbf{x}) + \hat{p}_\mathrm{tar}(\mathbf{x}|y)} \right)  = \log \left( \frac{2\hat{p}_\mathrm{tar}(\mathbf{x})}{q(\mathbf{x}) + \hat{p}_\mathrm{tar}(\mathbf{x})} \right).
\end{align}
\begin{IEEEproof}
    % \subsection{Proof of Lemma \ref{lem:2} }\label{subsec:prof_lem2}
Similar to the proof in Lemma \ref{lem:1}, if $\mathbf{x}\sim \hat{p}_\mathrm{tar}(y|\mathbf{x})$ for the target class $y$ and $1=\hat{p}_\mathrm{tar}(y|\mathbf{x})$, the following sequence of equalities show that Lemma \ref{lem:2} is true:
\begin{align}
        &\log \left( \frac{2\hat{p}_\mathrm{tar}(\mathbf{x}|y)}{q(\mathbf{x}) + \hat{p}_\mathrm{tar}(\mathbf{x}|y)} \right) \nonumber \\
        & = \log \left( \frac{2\hat{p}_\mathrm{tar}(\mathbf{x})\hat{p}_\mathrm{tar}(y|\mathbf{x})}{q(\mathbf{x}) + \hat{p}_\mathrm{tar}(\mathbf{x})\hat{p}_\mathrm{tar}(y|\mathbf{x})} \right) \\ 
        & = \log \left( \frac{2\hat{p}_\mathrm{tar}(\mathbf{x})}{q(\mathbf{x}) + \hat{p}_\mathrm{tar}(\mathbf{x})} \right). \nonumber
\end{align}

\end{IEEEproof}
\end{lemma}

\begin{theorem}\label{thm:1}
With \cref{assp:ineq_during_training}, the following inequality holds:
\if1\doublecolumn
\begin{equation}
\begin{split}
    & D_{\mathrm{JS}}\left(q(\mathbf{x}) \Vert \hat{p}_{\mathrm{tar}}(\mathbf{x}|y)\right) \label{eq:obj_fn_3} \\  
    & \le D_{\mathrm{JS}} \big( q(\mathbf{x}) \Vert \hat{p}_\mathrm{tar}(\mathbf{x}) \big) - \frac{1}{2}\mathbb{E}_{q(\mathbf{x})}\left[  \log\left(\hat{p}_\mathrm{tar}(y|\mathbf{x})\right) \right]. 
\end{split}
\end{equation}
\else
\begin{equation}
\begin{split}
    & D_{\mathrm{JS}}\left(q(\mathbf{x}) \Vert \hat{p}_{\mathrm{tar}}(\mathbf{x}|y)\right) \label{eq:obj_fn_3} \le D_{\mathrm{JS}} \big( q(\mathbf{x}) \Vert \hat{p}_\mathrm{tar}(\mathbf{x}) \big) - \frac{1}{2}\mathbb{E}_{q(\mathbf{x})}\left[  \log\left(\hat{p}_\mathrm{tar}(y|\mathbf{x})\right) \right]. 
\end{split}
\end{equation}
\fi
\begin{proof}
From Lemmas \ref{lem:1} and \ref{lem:2}, we can show that the following series of equations hold:
\begin{align}
    & D_{\mathrm{JS}}\left(q(\mathbf{x}) \Vert \hat{p}_{\mathrm{tar}}(\mathbf{x}|y)\right) \nonumber \\ 
    &= \frac{1}{2}D_{\mathrm{KL}}\left( q(\mathbf{x}) \Vert \frac{q(\mathbf{x}) + \hat{p}_{\mathrm{tar}}(\mathbf{x}|y)}{2} \right) \nonumber \\ 
         & ~~~~ + \frac{1}{2}D_{\mathrm{KL}}\left( \hat{p}_\mathrm{tar}(\mathbf{x}|y) \Vert \frac{q(\mathbf{x}) + \hat{p}_{\mathrm{tar}}(\mathbf{x}|y)}{2} \right) \nonumber \\ 
    &= \frac{1}{2} \mathbb{E}_{q(\mathbf{x})} \left[\log \left( \frac{2q(\mathbf{x})}{q(\mathbf{x}) + \hat{p}_\mathrm{tar}(\mathbf{x}|y))} \right)\right] \nonumber \\ 
        & ~~~~ + \frac{1}{2} \mathbb{E}_{\hat{p}_\mathrm{tar}(\mathbf{x}|y)} \left[\log \left( \frac{2\hat{p}_\mathrm{tar}(\mathbf{x}|y)}{q(\mathbf{x}) + \hat{p}_\mathrm{tar}(\mathbf{x}|y)} \right)\right] \\
    & \overset{(a)}{\le}  \frac{1}{2} \mathbb{E}_{q(\mathbf{x})} \left[  \log \left( \frac{2q(\mathbf{x})}{q(\mathbf{x}) + \hat{p}_\mathrm{tar}(\mathbf{x})} \right) \right] \nonumber \\ 
        & ~~~~~ + \frac{1}{2} \mathbb{E}_{\hat{p}_\mathrm{tar}(\mathbf{x})} \left[  \log \left( \frac{2\hat{p}_\mathrm{tar}(\mathbf{x})}{q(\mathbf{x}) + \hat{p}_\mathrm{tar}(\mathbf{x})} \right) \right] \nonumber \\ 
        & ~~~~~ -\frac{1}{2}\mathbb{E}_{q(\mathbf{x})}\left[  \log\left(\hat{p}_\mathrm{tar}(y|\mathbf{x})\right) \right] \nonumber \\
    &= D_{\mathrm{JS}} \big( q(\mathbf{x}) \Vert \hat{p}_\mathrm{tar}(\mathbf{x}) \big) - \frac{1}{2}\mathbb{E}_{q(\mathbf{x})}\left[  \log\left(\hat{p}_\text{tar}(y|\mathbf{x})\right) \right], \nonumber 
\end{align}
where inequality (a) holds since Lemmas \ref{lem:1} and \ref{lem:2} are true.
% The proof is shown in Appendix \ref{subsec:proof_thm1}.
\end{proof}
\end{theorem}

Theorem \ref{thm:1} delineates the upper bound of our objective function, conforming to \cref{assp:ineq_during_training}. 
% Since the target dataset is inaccessible, we neglect the last term in \eqref{eq:obj_fn_3}.\footnote{We note that the neglected term is zero if $\hat{p}_\mathrm{tar}(y|\mathbf{x}) = q(y|\mathbf{x})$, and the second term aims to approximate  $\hat{p}_\mathrm{tar}(y|\mathbf{x})$ to $q(y|\mathbf{x})$.}
Here, we reformulate the MI optimization problem as 
\begin{equation}\label{eq:obj_fn_2}
\begin{split}
     \arg\min_{q\sim Q_\mathbf{x}}D_\text{JS}(q(\mathbf{x}) & \Vert \hat{p}_\mathrm{tar}(\mathbf{x}))-\lambda \mathbb{E}_{q(\mathbf{x})} [\log(\hat{p}_\mathrm{tar}(y|\mathbf{\mathbf{x}}))],
\end{split}
\end{equation}
where a weight $\lambda\ge1/2$ is added to the attacker loss to control the posterior probability. As discussed in \cite{wang2021variational}, the power posterior \cite{knoblauch2019generalized} let us control the effect of posterior probability in bayes prior, \ie $q^*_\lambda(\mathbf{x})\propto p_{\text{aux}}(\mathbf{x})\hat{p}_\text{tar}^\lambda(y|\mathbf{x})$.

\begin{figure*}[t]
\centering    \includegraphics[width=1.\linewidth]{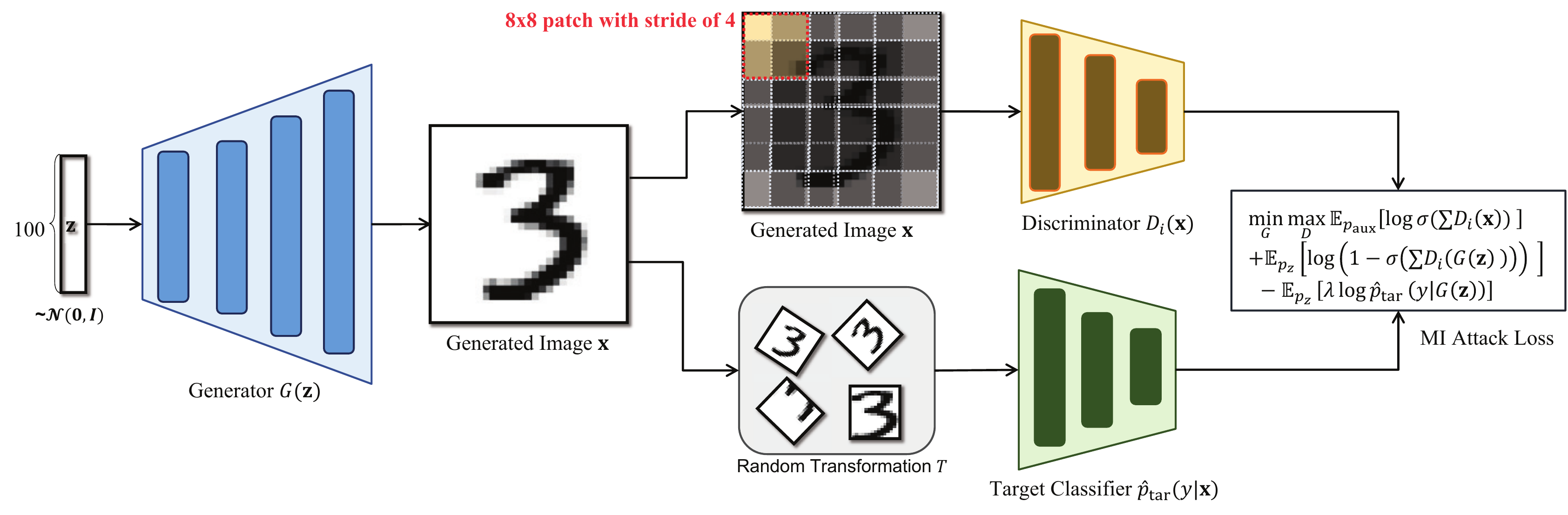}
% \vspace{-10pt}
    \caption{
    \textbf{Illustration of the Patch-MI attack method. }
    For the generator, we simply reuse the standard DCGAN structure for image generation. 
    For the patch-wise discriminator, we commence by applying a \texttt{Conv2d} layer, utilizing a filter whose size corresponds to the patch size and a stride size determined by the subtraction of overlapped pixels from the patch size. Subsequently, 1x1 convolution layers calculate $D_i$.
    Furthermore, the generated images are subjected to random transformation before being forwarded to the target classifier.
    % By consolidating all results and minimizing the MI attack loss, we successfully facilitate the MI attack.
    }
    \label{fig:GENDISCRIM}
\end{figure*}

\paragraph*{Patch-MI loss function}
At the beginning of this section, we assume that the attacker can access an auxiliary dataset that helps the attacker to generate photo-realistic images.
Now, in Equation \eqref{eq:obj_fn_2}, we can change the JS divergence term into a GAN objective function by defining a discriminator $\sigma(D(\mathbf{x})) = \frac{\hat{p}_\mathrm{tar}(\mathbf{x})}{\hat{p}_\mathrm{tar}(\mathbf{x}) + q(\mathbf{x})}$, where $\sigma(a) = 1/(1+\exp(-a))$, \ie $D(\mathbf{x}) = \log\frac{\hat{p}_\mathrm{tar}(\mathbf{x})}{q(\mathbf{x})}$.  
In addition, we define a generator $G(\mathbf{z})$ representing $q(\mathbf{x})$. We note that $q(\mathbf{x})$ can be sampled by $G(\mathbf{z})$, $\mathbf{z}\sim \mathcal{N}(\mathbf{0}, \mathbf{I})$. 
Then, the MI attack problem can be formulated to generate images that 1) imitate the target dataset distribution and 2) maximize the target posterior probability, as follows: 
\if1\doublecolumn
\begin{equation}
\begin{split}
     &\arg\min_{G} \max_{D} \underset{\mathbf{x}\sim p_\mathrm{tar}}{\mathbb{E}}\left[\log \sigma(D(\mathbf{x}))\right] \\ 
     & + 
     \mathbb{E}_{\mathbf{z}}\left[ \log\big(1- \sigma(D(G(\mathbf{z}))\big)-\lambda \log\hat{p}_\mathrm{tar} (y|G(\mathbf{z})) \right]. 
     % \nonumber
\end{split}
\end{equation}
\else
\begin{equation}
\begin{split}
     &\arg\min_{G} \max_{D} \underset{\mathbf{x}\sim p_\mathrm{tar}}{\mathbb{E}}\left[\log \sigma(D(\mathbf{x}))\right]  + 
     \mathbb{E}_{\mathbf{z}}\left[ \log\big(1- \sigma(D(G(\mathbf{z}))\big)-\lambda \log\hat{p}_\mathrm{tar} (y|G(\mathbf{z})) \right]. 
     % \nonumber
\end{split}
\end{equation}
\fi

As in previous studies \cite{Yang_19_GANMI, yang2019adversarial, wang2021variational}, our next step is to approximate the distribution of the target dataset into the auxiliary dataset. 
However, instead of the distribution of the whole image, we assume that the distributions of image patches from the target and auxiliary datasets are identical. The following assumption enables us to design our patch-based MI attack. 

\begin{assumption}[Partially similar auxiliary dataset]\label{assp:patch}
Let us consider a sequence of overlapped patches $x_1, \ldots, x_N$ of an image $\mathbf{x}$ and the neighboring pixels of the $i$-th patch $x_{-i}$. It is assumed that the patches are conditionally independent, which means that $\prod_{i=1}^N p_\mathrm{tar}(x_{i}|x_{-i}) = p_\mathrm{tar}(\mathbf{x})$.    
The distribution of the auxiliary dataset and the target dataset for the overlapped patches are approximately the same; that is, $p_\mathrm{tar}(x_i|x_{-i})\approx p_\mathrm{aux}(x_i|x_{-i})$. 
\end{assumption}

% \begin{proposition}\label{prop:ineq_proportion}
% Under Assumption 3, the JS divergence term can be decomposed for each patch, \ie, the following inequality holds:
% \begin{align}\label{eq:patch_obj}
% & D_\mathrm{JS}(q(\mathbf{x}) \Vert \hat{p}_\mathrm{tar}(\mathbf{x})) \le \sum_{i=1}^{N} D_\mathrm{JS}(q(x_i|x_{-i}) \Vert \hat{p}_\mathrm{tar}(x_i|x_{-i})) .
% \end{align}
% The proof is shown in  Appendix A.2.
% \end{proposition}

Under Assumption \ref{assp:patch}, by defining the discriminator for the $i$-th patch of an image $\mathbf{x}$ as $D_i$ the optimization of the discriminator can be decomposed to per-patch discriminators, as follows:
\begin{equation}
\begin{split}  
    D(\mathbf{x})&  =  \log\frac{\hat{p}_\mathrm{tar}(\mathbf{x})}{q(\mathbf{x})}= \sum_{i=1}^N\log\frac{\hat{p}_\mathrm{tar}(x_i|x_{-i})}{q(x_i|x_{-i})} =  \sum_{i=1}^{N}D_i(\mathbf{x}).
\end{split}
\end{equation}
% To solve the problem in  \eqref{eq:obj_fn_3}, we define a generator $G(\mathbf{z},y)$ representing $q(\mathbf{x}|y)$. We note that $q(\mathbf{x})$ can be sampled by $G(\mathbf{z},y)$, $y\sim \mathrm{Uniform}(0,1,...,C-1)$ and $z\sim \mathcal{N}(0, I)$. 
% For the discriminator, we define the discriminator has $N$ outputs, where $D_i(\mathbf{x})$ denotes the discriminator output for the $i$-th patch.
% By defining $D_i(\mathbf{x}) = \frac{\hat{p}_\mathrm{tar}(x_i|x_{-i})}{\hat{p}_\mathrm{tar}(x_i|x_{-i}) + q(x_i|x_{-i})} \approx \frac{p_\mathrm{aux}(x_i|x_{-i})}{p_\mathrm{aux}(x_i|x_{-i}) + q(x_i|x_{-i})}$, the problem in \eqref{eq:obj_fn_3} is rewritten as \eqref{eq:patch_obj} as 
Then, the Patch-MI loss function can be rewritten as 
\if1\doublecolumn
\begin{equation}\label{eq:obj_fn_4}
\begin{split}
    % &  \arg\min_{q\sim Q_\mathbf{x}}\sum_{i=1}^{N} D_\mathrm{JS}(q(x_i|x_{-i}) \Vert \hat{p}_\mathrm{tar}(x_i|x_{-i}))   -\lambda \mathbb{E}_{q(\mathbf{x},y)} [\log(\hat{p}_\mathrm{tar}(y|\mathbf{\mathbf{x}}))] \\
    &\arg\min_{G}  \max_{D} \mathbb{E}_{\mathbf{x}\sim p_\mathrm{aux}}\left[\log \sigma\left(\sum_{i=1}^N D_i(\mathbf{x})\right)\right] \\
    & ~+ \mathbb{E}_{\mathbf{z}}\Bigg[ \log\left(1- \sigma\left(\sum_{i=1}^N D_i(G(\mathbf{z}))\right)\right) \\ 
    & ~~~~~~~~~~~~~~~~-\lambda \log\hat{p}_\mathrm{tar} (y|G(\mathbf{z})) \Bigg].
    \end{split}
\end{equation}
\else
\begin{equation}\label{eq:obj_fn_4}
\begin{split}
    % &  \arg\min_{q\sim Q_\mathbf{x}}\sum_{i=1}^{N} D_\mathrm{JS}(q(x_i|x_{-i}) \Vert \hat{p}_\mathrm{tar}(x_i|x_{-i}))   -\lambda \mathbb{E}_{q(\mathbf{x},y)} [\log(\hat{p}_\mathrm{tar}(y|\mathbf{\mathbf{x}}))] \\
    &\arg\min_{G}  \max_{D} \mathbb{E}_{\mathbf{x}\sim p_\mathrm{aux}}\left[\log \sigma\left(\sum_{i=1}^N D_i(\mathbf{x})\right)\right] \\
    & ~~~~~~~~~~~~~~~~~~~~~+ \mathbb{E}_{\mathbf{z}}\Bigg[ \log\left(1- \sigma\left(\sum_{i=1}^N D_i(G(\mathbf{z}))\right)\right) -\lambda \log\hat{p}_\mathrm{tar} (y|G(\mathbf{z})) \Bigg].
    \end{split}
\end{equation}
\fi
% This approximation is followed by the fact that the optimal $D_i$ for given generator $G$ is 
% $\frac{p_\mathrm{aux}(x_i|x_{-i})}{p_\mathrm{aux}(x_i|x_{-i}) + q(x_i|x_{-i})}$.

\begin{proposition}\label{prop:infogan}
As similar to InfoGAN paper \cite{chen2016infogan}, the mutual information between the target classes $y$ and the generative images $G(\mathbf{z})$ is bounded by 
\if1\doublecolumn
\begin{equation}
\begin{split}
    &I(y ; G(\mathbf{z})) = H(y) - H(y|G(\mathbf{z})) \\
    & = H(y) + \mathbb{E}_{\mathbf{x}\sim G(\mathbf{z})}\left[ \mathbb{E}_{y\sim q(y|\mathbf{x})}[\log q(y|\mathbf{x})] \right]\\
    &= D_\mathrm{KL}\big( q(y|\mathbf{x}) \Vert \hat{p}_\mathrm{tar}(y|\mathbf{x})\big)  + H(y) \\ 
    & ~~~~~~+ \mathbb{E}_{\mathbf{x}\sim G(\mathbf{z})}\left[ \log \hat{p}_\text{tar}(y|\mathbf{x})\right]\\
    & \ge  \mathbb{E}_{\mathbf{x}\sim G(\mathbf{z})}\left[ \log \hat{p}_\text{tar}(y|\mathbf{x})\right].
\end{split}
\end{equation}
\else
\begin{equation}
\begin{split}
    &I(y ; G(\mathbf{z})) = H(y) - H(y|G(\mathbf{z})) \\
    & = H(y) + \mathbb{E}_{\mathbf{x}\sim G(\mathbf{z})}\left[ \mathbb{E}_{y\sim q(y|\mathbf{x})}[\log q(y|\mathbf{x})] \right]\\
    &= D_\mathrm{KL}\big( q(y|\mathbf{x}) \Vert \hat{p}_\mathrm{tar}(y|\mathbf{x})\big)  + H(y) + \mathbb{E}_{\mathbf{x}\sim G(\mathbf{z})}\left[ \log \hat{p}_\text{tar}(y|\mathbf{x})\right]\\
    & \ge  \mathbb{E}_{\mathbf{x}\sim G(\mathbf{z})}\left[ \log \hat{p}_\text{tar}(y|\mathbf{x})\right].
\end{split}
\end{equation}
\fi
\end{proposition}

Then, by using the result of Proposition \ref{prop:infogan}, our objective function is to learn generator $G(\mathbf{z})$ that imitates the auxiliary data distribution $p_\mathrm{aux}(x_i|x_{-i})$. While training, another objective is to maximize the mutual information between the target class $y$ and the generated images $G(\mathbf{z})$.
In other words, our method is analogous to replacing the categorical latent vector in InfoGAN with the target classifier.

\subsection{Patch Discriminator with Overlapped Patches  }

In \cref{fig:GENDISCRIM}, the schematic representation of the Patch-MI approach is depicted. The generator $G(\cdot)$ employs the standard DCGAN architecture \cite{radford2015unsupervised}, utilizing transposed convolution layers to create images from the random vector $\mathbf{z}\sim\mathcal{N}(\mathbf{0},\mathbf{I})$.
Contrary to earlier studies, our discriminator differentiates fake patches from realistic ones, rather than the entire image. In a manner akin to ViTGAN \cite{lee2021vitgan}, patches are overlapped to ensure photorealism, followed by embedding into a fixed-length vector using the \texttt{Conv2d} layer. Subsequently, the discriminators' outputs $D_i$ are obtained via $1\times1$ convolution.

\begin{figure}[t]
    \centering
    \subfloat[Patch-MI with patch size of 8 and stride of 4. ($\lambda=0$) \label{subfig:GAN_vs_Our_a}]{\includegraphics[width = 0.45\linewidth]{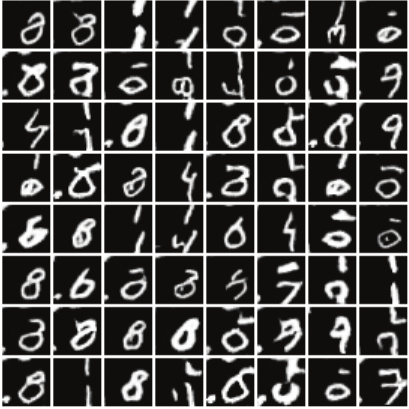}}
    \hspace{0.05\linewidth}
    \subfloat[Canonical GAN. \label{subfig:GAN_vs_Our_b}]{\includegraphics[width = 0.45\linewidth]{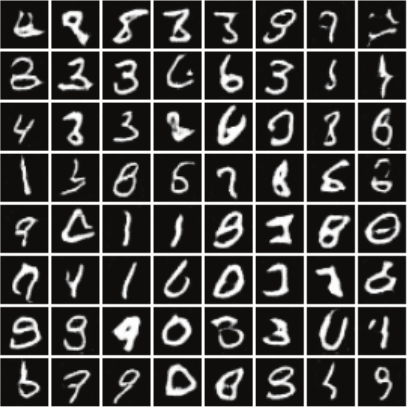}} 
    \caption{
    Examples of the generated images of our method and canonical GAN. Both generative models are trained with the \textbf{MNIST dataset}, \ie, handwritten digits from 0 to 9. 
    Within our approach, the patch and stride sizes of the discriminator are set at 8 and 4, respectively, given an image size of 32. 
    As evident from the figure, our technique is \textbf{capable of creating images that are not contained within the MNIST dataset}.
    }
    \label{fig:GAN_vs_Our}
\end{figure}
\begin{figure}[t]
    \centering
    \subfloat[\textbf{Experiment 1.} 
    % The target dataset is MNIST, and the auxiliary dataset is SERI95. 
    \label{subfig:exp1_setup}]{\includegraphics[width=\if1\doublecolumn 0.45 \else 0.3 \fi \linewidth]{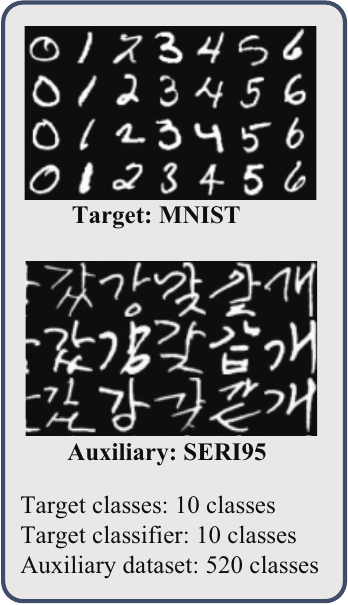} }
    \hspace{0.03\linewidth}
    \subfloat[\textbf{Experiment 2.} 
    % The target dataset is EMNIST-letter, and the auxiliary dataset is SERI95. 
    \label{subfig:exp2_setup}]{\includegraphics[width=\if1\doublecolumn 0.45 \else 0.3 \fi \linewidth]{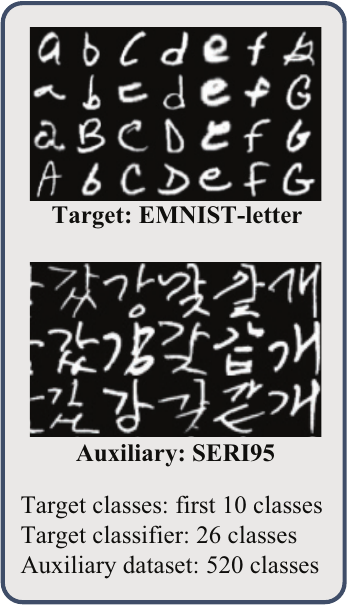} }
    \caption{Depiction of the target dataset and auxiliary dataset for two distinct experiments. In \protect\subref{subfig:exp1_setup}, the target dataset is designated as MNIST, and the auxiliary dataset is identified as SERI95. Analogously, the target dataset in \protect\subref{subfig:exp2_setup} is EMNIST-letter. 
    % In the second experiment, the focus is directed at the first 10 classes from a total of 26 classes.
    }
    \label{fig:Exp_setup}
\end{figure}

Figure \ref{fig:GAN_vs_Our} showcases images generated by both standard GAN and our approach with $\lambda=0$ for comparison. Notably, both methodologies are trained on the MNIST dataset, with our method displaying greater flexibility in image generation compared to the traditional GAN.

\subsection{Data Augmentation on Synthetic Images}
In \cite{pmlr-v162-struppek22a}, a data selection algorithm is implemented to devise a more general MI attack.
Under this approach, images undergo random transformations or augmentations, with selections being made based on the highest average confidence level over diverse transformations. 
A parallel intuition is evident in \cite{wang2022reconstructing}, where the incorporation of additional ensemble models is demonstrated to boost the efficacy of the MI attacker.
Mirroring these approaches, we utilize data augmentation to enable the generator $G$ to acquire more accurate images. 
Let us define the image augmentation function as $T$, which can be \texttt{Resize}, \texttt{RandomCrop}, \texttt{RandomRotation}, \etc. 
Subsequently, we task the image generator $G$ with maximizing $\log \hat{p}_\text{tar} (T\circ G(\mathbf{z}))$ instead of $\log \hat{p}_{\text{tar}}(G(\mathbf{z}))$, as shown in \cref{fig:GENDISCRIM}.

% \section{Experimental Details}\label{sec:exp_details}

% In this section, we detail the experimental setup for our study, describing the datasets, model architectures, training procedures, and specific configurations used in our experiments.

\paragraph*{Additional loss function for MNIST and EMNIST experiments}

In addition to our loss formulation in \eqref{eq:obj_fn_4}, we add an additional loss function used in \cite{Yin_2020_CVPR}. 
The added loss function aims to make the generated data follows the mean and standard deviation of the batch normalization layers in the target neural network.
By adding this, the performance of our method is enhanced.

% \subsection{Experiment Setup}

\section{Main Experimental Results}\label{sec:experiments}

In this section, we evaluate the proposed MI attack (Patch-MI) and show its superiority by comparing it to existing MI attacks.

\subsection{Experimental Details}

\paragraph*{Hardware} Our experiments are conducted at a workstation with  12th Gen Intel(R) Core(TM) i9-12900K 16-Core Processor CPU @ 5.20GHz and one NVIDIA Geforce RTX 3090 GPU.

\paragraph*{Experimental scenario}
In our experiments, we employ three experiments of the following (target, auxiliary) dataset pairs to evaluate MI attack methods on the cross-domain dataset scenarios: 1) MNIST \cite{lecun2010mnist} and SERI95 \cite{park2013evaluation} and 2) EMNIST-letter \cite{cohen2017emnist} and SERI95.
For the MNIST and EMNIST-letter datasets, our objective focuses on the reconstruction of image shapes within the target datasets, leading us to select another auxiliary dataset that excludes digits and alphabets. 
We cordially introduce SERI95, which encompasses 520 of the most commonly used Hanguel characters, each represented by approximately 1000 samples/class. The design of the experiments is illustrated in \cref{fig:Exp_setup}. 

Aside from this, we conduct additional experimental results for the CIFAR10 experiment, where we exclude images in the CIFAR100 dataset, whose labels correspond to the CIFAR10 dataset. 
For the additional experiments, please refer to Section \ref{sec:cifar10_appendix}.
% Within the main manuscript, the discussion is confined to results for gray-scale images, particularly those related to shape datasets. 
% For findings related to the color CIFAR10 dataset and further experimental details, please refer to Appendix \ref{sec:cifar10_appendix}.

\begin{table*}
  \caption{Evaluation results for our MI method on MNIST and EMNIST datasets, where the auxiliary dataset is SERI95.}
  \vskip 0.15in
  \label{table:overall_eval}
  \centering
\adjustbox{width=1\linewidth}{
    \begin{tabular}{llcccccccc}
    \toprule
    \textbf{Dataset} & \textbf{Methods}  & \textbf{Acc@1 $\uparrow$} & \textbf{Acc@5 $\uparrow$} & \textbf{Confidence $\uparrow$} & \textbf{Precision $\uparrow$} & \textbf{Coverage $\uparrow$} & \textbf{Density $\uparrow$} & \textbf{FID $\downarrow$} \\
    \midrule
    MNIST& BMI \cite{Fredrikson14}& 47.54\%& 91.67\% & 45.74\% & 0.0000& 0.0000& 0.0000& 407.7958\\
    & GMI \cite{Yang_19_GANMI}& 75.72\% & 99.13\% & 67.74\% & \underline{0.0115} & \underline{0.0051} & \underline{0.0016} & \underline{155.1654} \\
    & VMI\cite{wang2021variational}& \underline{94.25\%} & \underline{99.64\%} & \underline{89.27\%} & 0.0000& 0.0000& 0.0000& 204.4829\\
    & \textbf{Patch-MI (ours)} &\textbf{\underline{99.82\%}}& \textbf{\underline{100.00\%}} & \textbf{\underline{98.23\%}}& \textbf{\underline{0.0316}}&  \textbf{\underline{0.0240}}& \textbf{\underline{0.0131}}& \textbf{\underline{82.2328}}\\
    \cmidrule(r){1-9}
    EMNIST& BMI\cite{Fredrikson14}& 18.09\% & 74.36\%& 18.35\%& 0.0000& 0.0000& 0.0000& 426.1645 \\
    & GMI \cite{Yang_19_GANMI}& 54.20\%& 91.53\%& 46.75\%& \underline{0.0095}& \underline{0.0051}& \underline{0.0035}& \underline{140.8377} \\
    & VMI\cite{wang2021variational}& \underline{90.95\%}& \underline{98.99\%}& \underline{83.86\%}& 0.0005&  0.0002& 0.0004& 161.0970\\
    & \textbf{Patch-MI (ours)} &\textbf{\underline{98.20\%}}& \textbf{\underline{100.00\%}} & \textbf{\underline{93.62\%}}& \textbf{\underline{0.0153}}&  \textbf{\underline{0.0154}}& \textbf{\underline{0.0042}}& \textbf{\underline{84.6928}}\\
    \bottomrule
    \multicolumn{9}{l}{*Best: \underline{\textbf{bold and underline}}, second-best: \underline{underline}.}
    \end{tabular}
}
\end{table*}

\paragraph*{Baseline methods} 
In our benchmarking, we compare our methodology with 1) BMI \cite{Fredrikson14}, 2) GMI \cite{Yang_19_GANMI}, and 3) VMI \cite{wang2021variational}, each employing a common generator trained with DCGAN \cite{radford2015unsupervised}. Contemporary literature has manifested a prevailing trend focused on executing MI attacks at elevated resolutions \cite{wang2022reconstructing, pmlr-v162-struppek22a, han2023reinforcement}. Within these methodologies, auxiliary datasets used are those that either share classes with the target dataset or possess a structural resemblance. 
In contrast, our objective is to explore the situation that the target and auxiliary datasets have no overlapping classes. 
Thus, we orchestrates experiments in an unprecedented scenario wherein the distribution between the target and auxiliary datasets diverges markedly. 
In line with this, as foundational references for these investigations, we engage in comparative analysis with the elemental methodologies in low resolution, namely BMI, GMI, and VMI.

\paragraph*{Evaluation metrics}
For our evaluation, we train an evaluation classifier model with deeper layers, in which other configurations are set the same as the target classifier with different random seeds. 
Then, we assess the top-1 accuracy (Acc@1), top-5 accuracy (Acc@5), and confidence level for the target class, employing the evaluation classifier models. 
Subsequently, we leverage the Fr\'echet Inception Distance (FID) score \cite{heusel2017gans}, a well-known metric for measuring generated image qualities. 
This score means the distance between the feature vectors of images from the target dataset and the generated dataset, where extraction is performed using an Inception-v3 model \cite{xia2017inception} tuned on ImageNet \cite{Deng_jia_ImageNet}. 
The FID score corresponds to a heightened similarity between the two datasets; however, a per-class evaluation metric is still required. 
Thus, we utilize metrics such as improved precision, density and coverage \cite{kynkaanniemi2019improved,naeem2020reliable} on an individual label, to assess the diversity of the images, by following \cite{pmlr-v162-struppek22a}.

\subsection{Target Classifiers}
% We trained the ResNet-18 model \cite{he2016deep} for the target classifier and the DLA-34 model \cite{yu2018deep} for the evaluation classifier. 
% After training, the target classifiers have top-1 accuracy of 99.44\% for MNIST and  95.60\% for EMNIST-letter, respectively.
% Similarly, the evaluation classifiers have top-1 accuracy of 99.58\% for MNIST and  95.30\% for EMNIST-letter, respectively.

\paragraph*{MNIST}
We trained ResNet-18 as the target classifier and DLA-34 as the evaluation classifier. 
Both models are trained using the stochastic gradient descent (SGD) optimizer, with an initial learning rate of 2.5e-2, momentum of 0.9, and weight decay of 5.0e-4. The images are normalized with $\mu=\sigma=0.5$. The batch size is 256. 
The models are trained for 20 epochs.
The learning rate is reduced by a factor of 0.1 at 10, 15, and 18 epochs. After training, the validation accuracy is 99.44\% on the target classifier and 99.58\% on the evaluation classifier, respectively. 

% \paragraph*{MNIST}
% For the MNIST dataset, we employed the ResNet-18 model \cite{he2016deep} as the target classifier and the DLA-34 model \cite{yu2018deep} for evaluation. Both models were trained using SGD optimizer with specific parameters including an initial learning rate of $2.5\cdot 10^{-2}$, momentum of 0.9, and weight decay of $5.0\cdot 10^{-4}$. 
% The models are trained for 20 epochs.
% The images are normalized with $\mu=\sigma=0.5$. The batch size is 256. 
% The learning rate is reduced by a factor of 0.1 at the 10-th, 15-th, and 18-th epochs. After training, the validation accuracy is 99.44\% on the target classifier and 99.58\% on the evaluation classifier, respectively. 

\paragraph*{EMNIST-letter}
For the EMNIST-letter dataset, the training setup was similar to MNIST, with adjustments in the maximum epochs and learning rate schedule.
All the setups are the same except for maximum epochs of 30. Also, the learning rate is reduced by a factor of 0.1 at the 15-th, 22-nd, and 27-th epochs. After training is done, the validation accuracy is 95.60\% and 95.30\%, respectively.

% We note that the target classifiers have top-1 accuracy of 99.44\% for MNIST and  95.60\% for EMNIST-letter, respectively.
% Also, the evaluation classifiers have top-1 accuracy of 99.58\% for MNIST and  95.30\% for EMNIST-letter, respectively.

\subsection{MI Attacker}

\paragraph*{Generator and discriminator}
For the image generator and discriminator, we employ the canonical DCGAN \cite{radford2015unsupervised} model, which follows the standard DCGAN architecture~\cite{radford2015unsupervised}, with specifics on the latent vector, convolution layers, and activation functions. 
The latent vector $\mathbf{z}$ is forwarded into three consecutive transposed convolution layers with kernel size of 4, stride size of 2, and padding size of 1, in which rectified linear unit (ReLU) activation is applied except for the last layer. 
Similarly, the discriminator setup, including the embedding layer and convolution layers, is detailed, emphasizing their role in distinguishing real and fake patches.
In the discriminator, the \texttt{Conv2d} layer first embeds the image patches into vectors, with patch size of 8, stride size of 4, and padding size of 0.
Then, three 1x1 convolution layers are placed after the embedding convolution layer for distinguish real patches from fake patches. 
% , except for the modification in discriminator depicted in \cref{fig:GENDISCRIM}.
% , where the patch size and stride size are 8 and 4, respectively.
In the target posterior term in the loss function \eqref{eq:obj_fn_4}, we use $\lambda=30$.
We use Adam optimizer \cite{kingma2014adam} in our method, where the learning rate is fixed at $1.0\cdot 10^{-3}$. We trained 30 epochs for the target attacker.

\paragraph*{Output augmentation} 
In our method, the generated images $G(\mathbf{z})$ are randomly transformed before forwarding to the target classifier.
For experiments 1 and 2, the image transformer consists of 1) random rotation of [-20, 20] degrees, 2) random resized crop of 32x32 images with scales [0.85, 1.0] and ratio [0.9, 1.1].
Let us define the transform function as $T$. Then, the classifier output is defined by
\begin{equation}
    \frac{1}{2}\left( \hat{p}_\text{tar}(y|\mathbf{x}) + \hat{p}_\text{tar}(y|T(\mathbf{x})) \right).
\end{equation}

\paragraph*{GAN label smoothing}
In our experiments, the basic GAN loss function often diverges.
To address issues of divergence in the basic GAN loss function, we implemented one-sided label smoothing and explored the use of LS-GAN \cite{mao2017least}. 
Specifically, we add a one-sided label smooth by reducing the true label from 1.0 to 0.6. 
This assumes that 60\% of the auxiliary patches are false; thus, the generator no longer needs to learn all the auxiliary patches. 
In our experiments, adding a label smoothing trick helps convergence.

% We kindly note that the further experimental details are available in \cref{sec:exp_details}.

\begin{table*}
  \caption{Ablation study on our method with and without patch-wise discriminator and random transformation, on EMNIST-letter dataset. }
  \vskip 0.15in
  \label{tab:ablation_study}
  \centering
  \adjustbox{width=1\linewidth }{
  \begin{tabular}{ccccccccc}
    \toprule
    Patch size    & Trans. & \textbf{Acc@1 $\uparrow$} & \textbf{Acc@5 $\uparrow$} & \textbf{Confidence $\uparrow$} & \textbf{Precision $\uparrow$} & \textbf{Coverage $\uparrow$} & \textbf{Density $\uparrow$} & \textbf{FID $\downarrow$}  \\
    \midrule
    32 (full) &  \xmarkg & 69.82\% & 95.42\% & 59.56\% & 0.0000 &  0.0000 & 0.0000 & 229.5503 \\
    32 (full) & \cmark & 90.69\% & 99.71\% & 83.67\% & 0.0003 &  0.0002 & 0.0003 & 202.6455 \\
    8 & \xmarkg
    & 91.33\% & 99.67\% & 81.81\% & 0.0027 &  0.0015 & 0.0010 & 171.1700 \\
    \midrule
    8 & \cmark &\textbf{\underline{98.20\%}}& \textbf{\underline{100.00\%}} & \textbf{\underline{93.62\%}}& \textbf{\underline{0.0153}}&  \textbf{\underline{0.0154}}& \textbf{\underline{0.0042}}& \textbf{\underline{84.6928}}\\
    \bottomrule
    \multicolumn{9}{l}{*Best: \underline{\small\textbf{bold and underline}}.}
  \end{tabular}
  }
\end{table*}

\begin{figure*}[t]
\centering
    \subfloat[Experiment 1.\label{subfig:visualization_mnist}]{\includegraphics[width=0.49\linewidth]{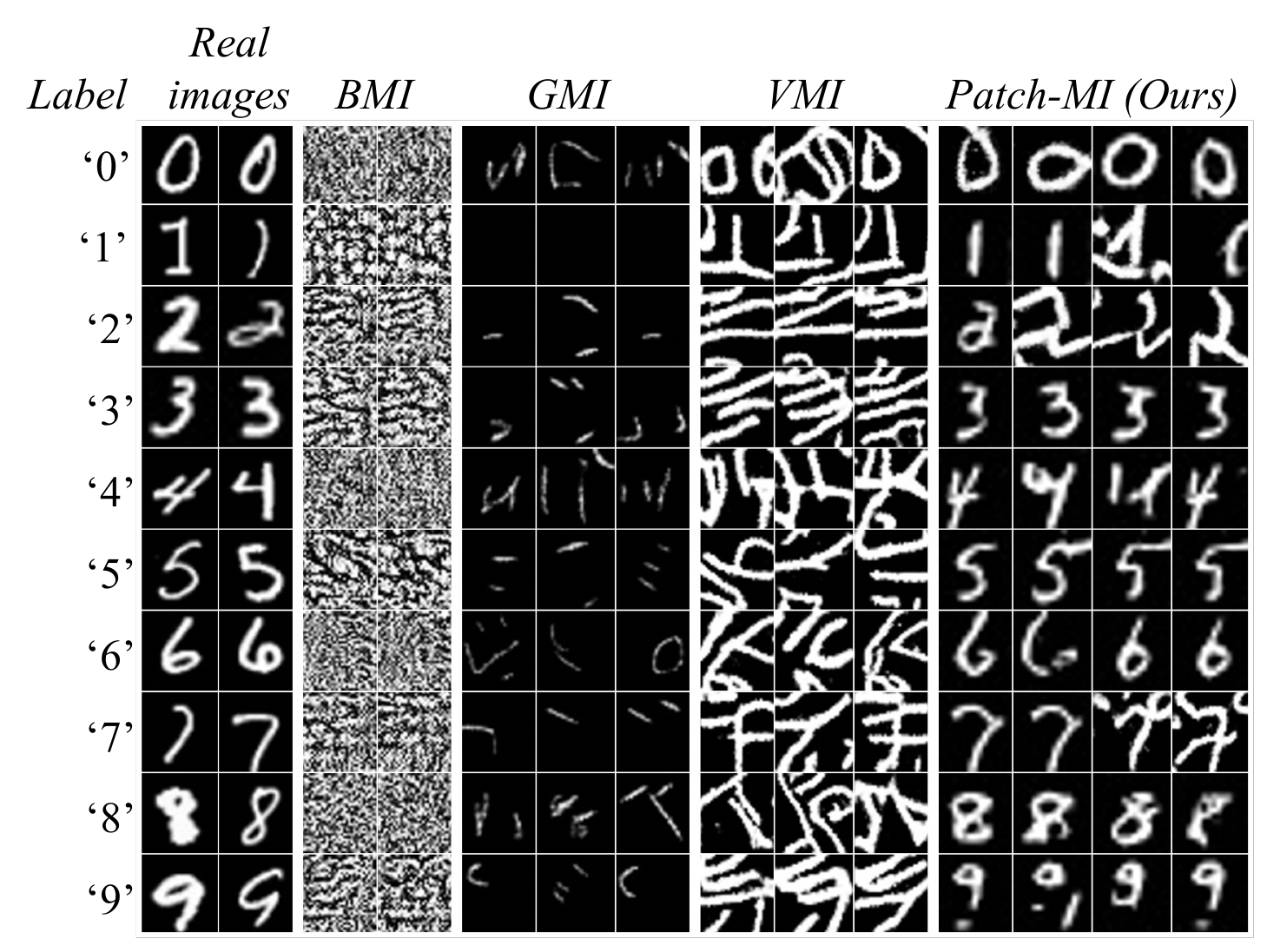} }
    \subfloat[Experiment 2. \label{subfig:visualization_emnist}]{\includegraphics[width=0.49\linewidth]{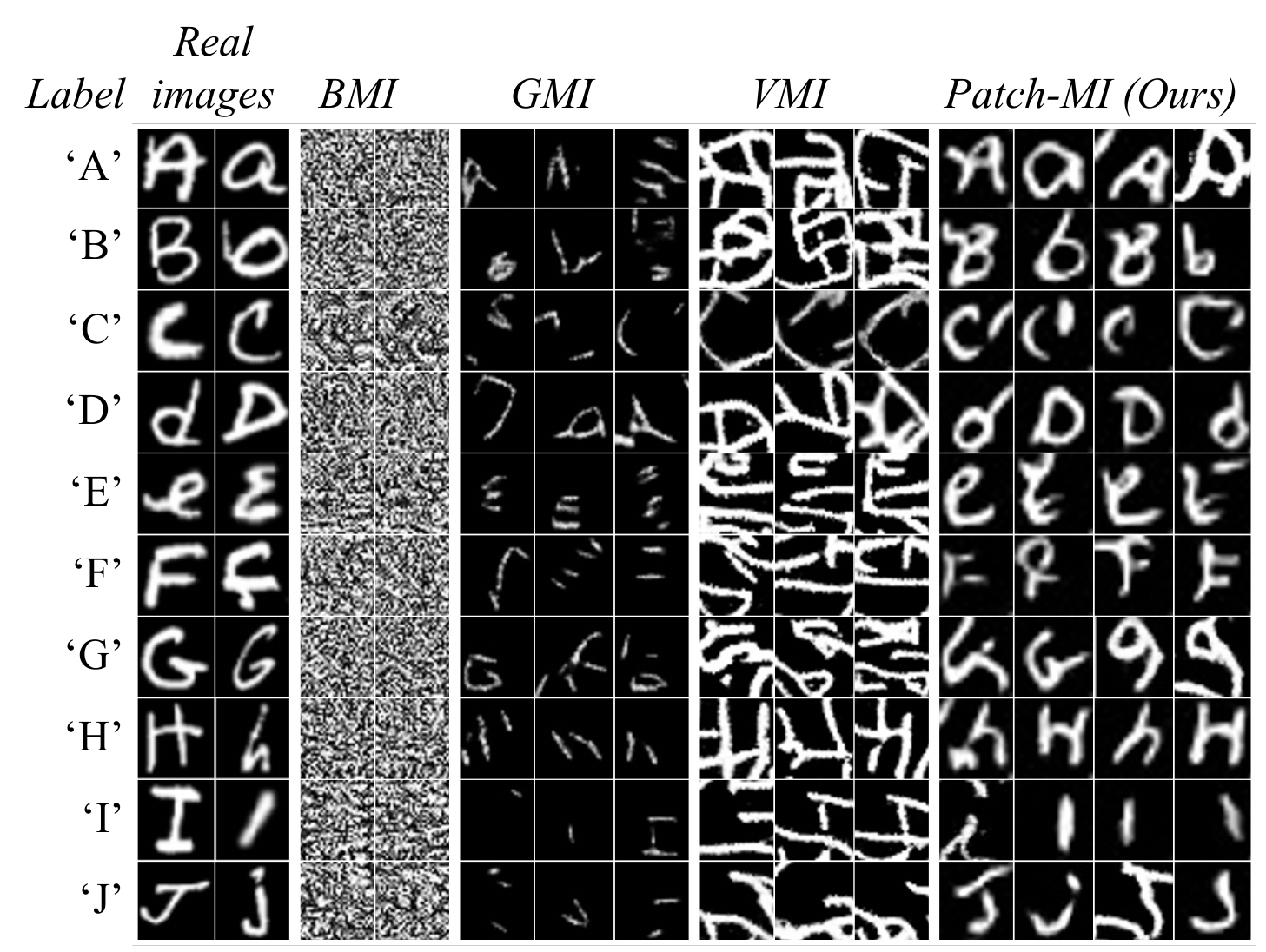}}
    % \centering
    % \includegraphics[width=1\linewidth]{Figure/MNIST_result-eps-converted-to.pdf}
    \caption{
    Visualization of the randomly chosen outputs for various MI attack methods on the experiments 1 and 2. 
    }
    \label{fig:Visualization}
\end{figure*}

\subsection{Comparison with Existing MI Attacks}

To evaluate the efficacy of our proposed method, we first conduct a comparative analysis against established baseline methods under the experimental settings of experiments 1 and 2.
In \cref{table:overall_eval}, we present the numerical results for these experiments, detailing the performance of four methodologies across various metrics.

\paragraph*{Accuracy metrics}
In the comparative analysis presented in the table, our proposed method consistently outperforms the baseline methods across all accuracy-related metrics (Acc@1, Acc@5, and Confidence level) in both experiments.
Specifically, in experiment 1 using the MNIST dataset, our method shows a significant enhancement in top-1 attack accuracy, improving by 5\%p, nearly reaching 100\%. 
In experiment 2, utilizing the EMNIST dataset, the performance gain is even more pronounced. 
Here, we observe a 9\%p improvement in top-1 accuracy and a 14\%p increase in the confidence level.
These results validate our hypothesis that the patch-based MI reconstruction and random transformation techniques significantly contribute to the enhanced accuracy of our attack method. 
This is particularly evident in the generalization capabilities observed in both MNIST and EMNIST experiments.

\paragraph*{Quality metrics}
Here, we compare the quality metrics (precision, coverage, density, and FID).
A notable finding from our analysis is that our proposed method consistently surpasses both BMI and VMI approaches across all these quality-related metrics.
However, it is interesting to note that the GMI method shows comparable results to our method in terms of quality metrics. 
Despite this similarity, GMI significantly lags behind in accuracy-related metrics.
This suggests that GMI potentially sacrifices attack accuracy in pursuit of more photo-realistic and diverse images.
While the identity loss weight in the GMI can be adjusted as per \eqref{eq:GMI} to strike a balance between accuracy and quality metrics, our method still outperforms GMI in terms of overall results.
This not only indicates the adaptability of GMI but more importantly, confirms the superiority of our approach, which achieves high-quality metrics without compromising attack accuracy.

These comparative results lead us to confidently assert the superiority of our method, which excels in both quality and accuracy metrics, addressing the limitations evident in other approaches.

\subsection{Ablation Study}

To further understand the impact of the patch-based reconstruction and ensemble optimization with random transformation, integral to our proposed method, we conducted an ablation study, the results of which are detailed in \cref{tab:ablation_study}.
Initially, when neither of these two features is applied, effectively using a full-size GAN without random transformation, we observed that the performance slightly surpasses GMI but falls short of VMI. 
We note that the proposed method without patch-based reconstruction and random transformation is quite equivalent to GMI except for directly training the GAN model to mimic the distribution of the target classifier.
Upon introducing patch-based reconstruction alone, with a patch size of 8, there's a noticeable improvement. The method begins to outperform both GMI and VMI, clearly demonstrating the significant role of patch-based reconstruction in enhancing MI attack performance.
With the incorporation of both patch-based reconstruction and random transformation, as previously detailed, the proposed method not only outperforms GMI and VMI but does so significantly. This underscores the synergistic effect of these features in our method.

In summary, the results from our ablation study underscore the nuanced interplay and importance of these key features in elevating the efficacy of our proposed MI attack method.

% \begin{figure}[t]
%     \centering
%     \includegraphics[width=1\linewidth]{Figure/EMNIST_result-eps-converted-to.pdf}
%     \caption{
%     Graphical Results of our method and other MI attack methods. 
%     (a) 
%     (b)
%     }
%     \label{fig:Visualization}
% \end{figure}

\subsection{Graphical Results}

Previously, we demonstrate the quantitative comparison of the proposed method and baseline methods.
Here, we now present graphical results to visually compare the MI attack methods on MNIST and EMNIST datasets, using SERI95 as the auxiliary dataset. 
In \cref{fig:Visualization}, sub-figures \protect\subref{subfig:visualization_mnist} and \protect\subref{subfig:visualization_emnist} offer detailed visualizations for each class of the target dataset
For a comprehensive comparison, we also include examples of original images from the target dataset alongside the MI method outputs.

The BMI results tend to generate noisy images, lacking in photo-realism, as evident from its objective function \eqref{eq:BMI}. 
This absence of a photo-realism term results in images that are visually less appealing and less accurate.
In the case of GMI, while the images are somewhat related to the target classes, common features, such as those seen in the third image of class 'C', indicate a compromise. 
This aligns with our findings in \cref{table:overall_eval}, confirming GMI's tendency to prioritize image quality over accuracy.
On the other hand, VMI shows improvements over BMI and GMI in terms of generating more class-related images. However, it still lacks the capability to conduct a patch-level MI attack.

Our proposed method, in contrast, demonstrates clearer and more accurate output images. This superiority is attributed to its ability to conduct patch-level MI attacks, effectively enhancing both the quality and accuracy of the generated images.

% In \cref{fig:Visualization}, the subfigures \protect\subref{subfig:visualization_mnist} and \protect\subref{subfig:visualization_emnist} depict the visualizations of various MI methods applied to the MNIST and EMNIST-letter datasets, respectively. Overall, BMI tends to generate noisy images, and both GMI and VMI exhibit a decrease in structural similarity, which can be observed as a limitation due to their inability to learn at the patch level. In contrast, the proposed method is observed to produce images with structural similarity, even when attacking an EMNIST-letter and MNIST classifiers using Korean Hanguel data. This finding underlines the capability of the proposed approach to maintain structural coherence across diverse contexts, highlighting its potential efficacy and robustness.

\begin{figure}[!ht]
    \centering
    \includegraphics[width=\if1\doublecolumn 1 \else 0.6 \fi\linewidth]{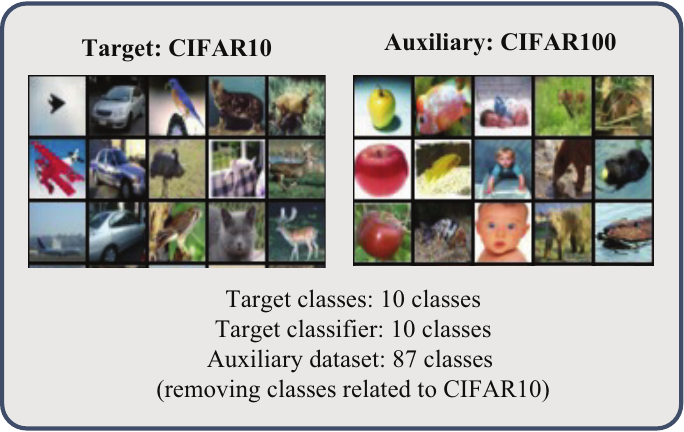}
    \caption{Depiction of the target dataset and auxiliary dataset in the experiment 3. 
  The target dataset is designated as CIFAR10, and the auxiliary dataset is identified as CIFAR100. Unfortunately, there are some \textbf{overlapped labeled data} such as (bus, pickup truck, street car, tractor, lawn mower, rocket, dolphin, whale, ray, train, shark, aquarium fish, and leopard) in the CIFAR100 dataset, \textbf{we remove them in our experiments.} }
    \label{fig:exp3_setup}
\end{figure}

\begin{table*}[t]
  \caption{ Evaluation results for our MI method on CIFAR10 datasets. Here, the dataset-wise metrics are written. }
\vskip 0.15in
  \label{tab:cifar10_acc}
  \centering
\adjustbox{width=0.85\linewidth}{
\begin{tabular}{lccccccc}
\toprule
\textbf{}  & \textbf{Acc@1 $\uparrow$} & \textbf{Acc@5 $\uparrow$} & \textbf{Confidence $\uparrow$} & \textbf{FID $\downarrow$} & \textbf{Precision $\uparrow$} & \textbf{Coverage $\uparrow$} & \textbf{Density $\uparrow$} \\
\midrule
BMI& 14.01\% & 56.61\% & 13.10\% & 427.8511& 0.0000&  0.0000& 0.0000\\
GMI& 12.36\% & 66.18\% & 12.91\% & 211.9117& 0.0795&  0.0717& 0.0211\\
VMI& \underline{83.14\%} & \underline{99.11\%} & \underline{76.73\%} & \textbf{\underline{48.4994}} & \underline{0.2085}&  \underline{0.4690}& \textbf{\underline{0.2638}}\\
Patch-MI (ours) & \textbf{\underline{96.47\%}} & \textbf{\underline{99.86\%}} & \textbf{\underline{91.66\%}} & \underline{93.7527} & \textbf{\underline{0.2543}}& \textbf{\underline{0.5190}}& \underline{0.1912} \\
\bottomrule
\multicolumn{5}{l}{\small*Best: \underline{\textbf{bold and underline}}, second-best: \underline{underline}.}
\end{tabular}
}
\vskip -0.1in
\end{table*}

\begin{figure*}
    \centering
    \includegraphics[width=.99\linewidth]{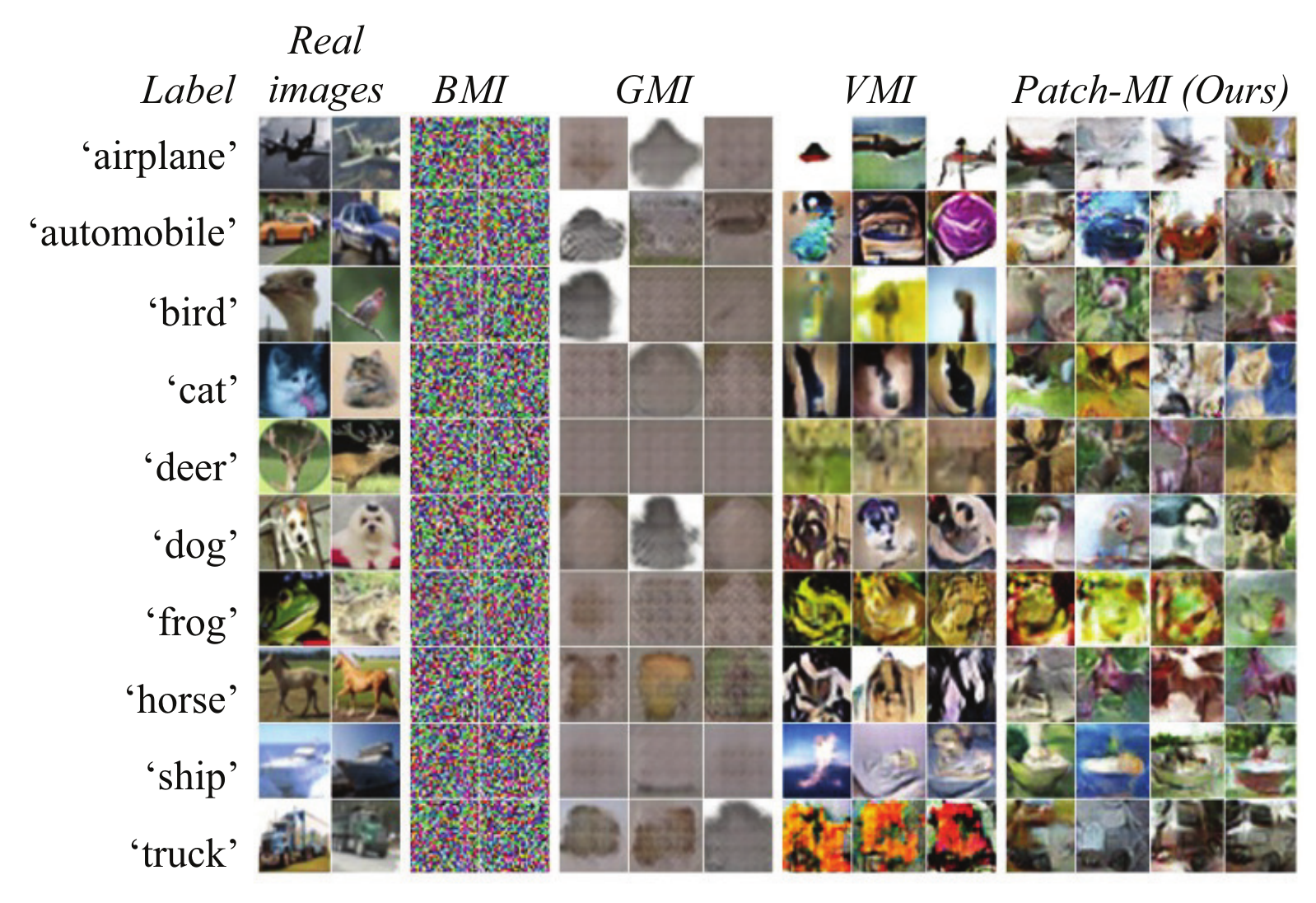}
    \caption{
    Visualization of the generated images via various MI attack methods on the CIFAR10 dataset. 
    }
    \label{fig:Visualization_cifar10}
\end{figure*}

\begin{table*}
\caption{Evaluation results of the proposed method and VMI, where the experiments are conducted to verify the tradeoff between the accuracy and other data statistical metrics. \label{table:various_lambda}}
\vskip 0.15in
\centering
\adjustbox{width=.8\linewidth}{
\begin{tabular}{lcccccccc}
\toprule
\textbf{}  & \textbf{Acc@1 $\uparrow$} & \textbf{Acc@5 $\uparrow$} & \textbf{Conf. $\uparrow$} & \textbf{FID $\downarrow$} & \textbf{Prec. $\uparrow$} & \textbf{Cov. $\uparrow$} & \textbf{Dens. $\uparrow$}\\
\midrule
PMI ($\lambda=50$) & \textbf{96.47\%} & \textbf{99.86\%} & \textbf{91.66\%} & 93.7527 &\textbf{0.2543}&  0.5190& 0.1912\\
% PMI ($\lambda=25$) & 94.29\% & 99.93\% & 88.70\% & 87.8770\\
PMI ($\lambda=10$) & 92.89\% & 99.66\% & 86.82\% & 85.0515 & 0.2512&  0.5086& 0.2153\\
PMI ($\lambda=5$)  & 89.01\% & 99.46\% & 82.78\% & 71.1711 & 0.2540 &  0.5216& 0.2370\\
PMI ($\lambda=2$)  & 81.54\% & 97.24\% & 76.44\% & 67.9040 & 0.2674 &  \textbf{0.5450}& 0.2224\\
\midrule
VMI ($w=10^{-1}$) & 78.66\% & 99.06\% & 72.65\% & 55.1312 & 0.2426&  0.4720& 0.2192 \\
% VMI ($w$=0.01)& 79.84\% & 98.97\% & 73.73\% & 46.1811\\
VMI ($w=10^{-3}$) & 83.14\% & 99.11\% & 76.73\% & \textbf{48.4994} & 0.2085&  0.4690& \textbf{0.2638}\\
% VMI ($w$=0.0001)& 83.14\% & 99.11\% & 76.73\% & 48.4994\\
% VMI ($w=1.0\cdot10^{-5}$)& 85.37\% & 99.47\% & 79.05\% & 49.9111\\
VMI ($w=10^{-7}$)& 83.99\% & 99.12\% & 76.45\% & 49.6053 & 0.2175&  0.5018& 0.2546\\
VMI ($w=10^{-13}$)& 65.37\% & 89.04\% & 60.59\% & 97.1975 & 0.1743&  0.3827& 0.1916\\
\bottomrule
\end{tabular}
}
\vskip -0.1in
\end{table*}

\begin{table}
\centering
\caption{Experimental results of PMI attacker for various classifiers: ResNeXt29~\cite{xie2017aggregated}, VGG11~\cite{simonyan2015deep}, VGG19, and MobileNet v2~\cite{sandler2019mobilenetv2}. We note that the classifiers are well-trained with the CIFAR 10 dataset. \label{tab:various_classifier}}
\vskip 0.15in
\adjustbox{width=\if1\doublecolumn 1 \else 0.6 \fi \linewidth}{
\begin{tabular}{lcccc}
\toprule
\textbf{}  & \textbf{Acc@1 $\uparrow$} & \textbf{Acc@5 $\uparrow$} & \textbf{Confidence $\uparrow$} & \textbf{FID $\downarrow$}  \\
\midrule
ResNet18 & 96.47\% & 99.86\% & 91.66\% & 93.7527 \\
ResNeXt29 (32x4d) & 93.70\% & 99.80\% & 88.68\% & 72.1683\\
VGG11& 93.81\% & 99.75\% & 88.99\% & 94.8382\\
VGG19& 99.09\% & 99.99\% & 95.43\% & 98.0052 \\
MobileNet V2& 99.28\% & 99.99\% & 96.19\% & 95.1164 \\
\bottomrule
\end{tabular}
}
\end{table}

\section{Additional Experiment: CIFAR10} \label{sec:cifar10_appendix}
In addition to experiments with black and white images targeting the MNIST and EMNIST datasets, we have conducted supplementary experiments on color images with the CIFAR10 dataset as the target, as depicted in \cref{fig:exp3_setup}.
% Detailed experimental setups and results will be discussed in the remainder of this section.

\paragraph*{Dataset}
 For the CIFAR100 dataset, \textit{the images with labels (bus, pickup truck, street car, tractor, lawn mower, rocket, dolphin, whale, ray, train, shark, aquarium fish, and leopard)} related with the CIFAR10 dataset is removed in our experiment.

\paragraph*{Created image augmentation}

In our method, the generated images $\mathbf{x}=G(\mathbf{z})$ is randomly transformed before forwarding to the target classifier.
For experiments 1 and 2, the image transformer consists of 1) padding 4 pixels, 2) random rotation of [-45, 45] degrees, and 3) random horizontal flip.
Let us define the transform function as $T'$. Then, the classifier output is defined by 
\begin{equation}
    \frac{1}{2}\left( \hat{p}_\text{tar}(y|\mathbf{x}) + \hat{p}_\text{tar}(y|T'(\mathbf{x})) \right).
\end{equation}

\subsection{Experimental Results}

In Table \ref{tab:cifar10_acc}, we evaluate various MI attack methods based on metrics encompassing the entire dataset and per-class metrics.
Similarly to the gray-scale dataset (MNIST and EMNIST-letter), the Patch-MI outperforms all other MI techniques in terms of accuracy. In contrast, the VMI exhibits better performance than our proposed approach concerning the FID score. However, in the per-class results, VMI and Patch-MI were found to have similar performance. 
That is, in the per-class statistics, VMI and Patch-MI have similar characteristics. 
The graphical results in \cref{fig:Visualization_cifar10} show the reason why the Patch-MI has a lower FID than the VMI:
the Patch-MI creates images similar to the target label by slightly sacrificing photo-realism, whereas VMI generates more diverse images that are distant from the target image but classified with lower confidence levels to the corresponding label.

Hence, it was observed that Patch-MI has a lower FID score than VMI due to the diversity of output images. Conversely, when viewed on a per-class basis, the statistical quality is similar between VMI and Patch-MI, and in terms of accuracy, Patch-MI overwhelms other methods.

\paragraph*{Further analysis with various attacker weights $\lambda$}

To address the conflicting metrics with accuracy: precision, coverage, and density, we conducted an experiment involving various values of attacker weight for the CIFAR 10 dataset. The numerical results are presented in Table \ref{table:various_lambda}.

As indicated in the table, our proposed method outperforms the VMI scheme, albeit with a slight sacrifice in attack accuracy. However, the accuracy of our method remains higher than that of VMI. More significantly, our method excels in precision and coverage metrics, and is comparable in terms of density. This demonstrates that our method is superior to VMI in aspects other than FID, which is a density-based metric. Our analysis suggests that precision, coverage, and density provide a more accurate assessment than FID.

\begin{itemize}
    \item \textbf{Tradeoff:} The table illustrates a fundamental tradeoff between quality-related metrics (precision, coverage, and density) and accuracy-related metrics (Acc@1, confidence). While our method exhibits superior performance in quality-related metrics, there is a noticeable balance between these metrics and the traditional accuracy-related metrics. This tradeoff highlights the complexity of optimizing for multiple metrics in machine learning models and underscores the necessity of considering a diverse range of evaluation criteria to comprehensively assess model performance.
\end{itemize}

\paragraph*{Different target classifiers}

In the previous results, we have experimented with a target classifier. Here, we execute our method to attack various classifiers: ResNeXt29, VGG11, VGG19, and MobileNet v2 with the CIFAR 10 dataset, where all the classifiers are well-trained with many epochs.
In Table \ref{tab:various_classifier}, we denote the accuracy, confidence level, and FID score of the image obtained via PMI attack. 
As shown in the table, the proposed method consistently has high accuracy and high confidence in all the classifiers, while performing FID scores less than 100.0.
These results note that our method can work generally on various target classifiers. 

\section{Discussion}
\label{sec:discussion}

In summary, our novel Patch-MI method represents a significant advancement in MI attack methodologies. 
Unlike existing works, which often rely on access to partially overlapped auxiliary datasets, Patch-MI leverages patch-level data statistics from any given auxiliary dataset to accurately mimic the target dataset distribution.
We formulate an optimization problem that effectively minimizes the JS divergence between target and generated images, offering a more nuanced and effective approach to MI attacks.
Our theoretical analysis provides a new probabilistic perspective on MI attacks, while our experimental results clearly demonstrate Patch-MI's superiority, particularly when dealing with non-overlapping target and auxiliary datasets.
Even though the auxiliary datasets given in our experiments do not overlap with the target dataset, the proposed method has 99\%++ attack accuracy, while outperforming the existing works' accuracies. 

\paragraph*{Positive effect: Against potential defensive methods}
The differentially private stochastic gradient descent (DP-SGD) method is a representative method to ensure data privacy of trained neural network models~\cite{dpsgd,Zhu2021-yr}.
Regarding potential defenses, we found that even against DP-SGD, Patch-MI maintains high attack accuracy. 
For instance, the Patch-MI's attack accuracy is higher than 99\% (if $\epsilon=4.0$), even though DP-SGD is applied to the target MNIST classifier. 
This indicates a need for the development of more robust defense mechanisms, thereby contributing to the overall enhancement of data privacy.

\paragraph*{Comparison with recent generative MI attacks}

Many previous studies ~\cite{Fredrikson14,Fredrikson15,Hidano17,knoblauch2019generalized,zhang2020secret,wang2021variational} have shown that well pre-trained GANs, such as Style-GAN, can generate images that are much cleaner and more realistic than our results (\cref{fig:Visualization}). However, the methods used in those studies imply that the target and auxiliary data classes are partially overlapped, or that the generative model itself is capable of generating images of those classes. In our work, similar to another work \cite{yu2024generator}, we perform MI attacks without an auxiliary dataset or a pre-trained GAN similar to the target dataset, and we observe that our method generates much cleaner images compared to the results of that work.

\paragraph*{Limitations}
Despite its success, the proposed method has limitations. 
Our experiments primarily focused on low-resolution datasets, and while we attempted a more challenging task with the CelebA classifiers using the CIFAR10 auxiliary dataset, the results showed lower attack accuracy ($\approx 30\%$).
This suggests the need for further refinement for high-resolution image tasks. 
Additionally, our method currently assumes a white-box attack setting. Exploring its adaptability to gray-box or black-box scenarios is an exciting avenue for future research.

% \appendices

% \section{Proof of Theorem 1 }
% \label{subsec:proof_thm1}
% \setcounter{theorem}{0}
% \begin{theorem}
% With \cref{assp:ineq_during_training}, the following inequality holds:
% \begin{align}
%     & D_{\mathrm{JS}}\left(q(\mathbf{x}) \Vert \hat{p}_{\mathrm{tar}}(\mathbf{x}|y)\right) \\
%     & ~~~~ 
%     \le  D_{\mathrm{JS}} \big( q(\mathbf{x}) \Vert \hat{p}_\mathrm{tar}(\mathbf{x}) \big) - \frac{1}{2}\mathbb{E}_{q(\mathbf{x})}\left[  \log\left(\hat{p}_\text{tar}(y|\mathbf{x})\right) \right]  \nonumber.
% \end{align}
% \end{theorem}

\bibliographystyle{IEEEtran}
\bibliography{main}

\end{document}